# CLUSTER APPROACH TO THE DOMAINS FORMATION

Leonid B. Litinskii

Centre of Optical-Neural Technologies

Scientific-Research Institute for System Investigates Russian Academy of Sciences

CONT, Vavilov str., 44/2, Moscow, 119333, Russia

litin@mail.ru

**Abstract.** As a rule, a quadratic functional depending on a great number of binary variables has a lot of local minima. One of approaches allowing one to find in averaged deeper local minima is aggregation of binary variables into larger blocks/domains. To minimize the functional one has to change the states of aggregated variables (domains). In the present publication we discuss methods of domains formation. It is shown that the best results are obtained when domains are formed by variables that are strongly connected with each other.

## 1. INTODUCTION

We discuss the problem of minimization of a quadratic functional depending on $N$ binary variables $s_i = \{\pm 1\}$:

$$E(\mathbf{s}) = -(\mathbf{Js},\mathbf{s}) = -\sum_{i,j=1}^{N} J_{ij} s_i s_j \xrightarrow[\mathbf{s}]{} \min. \qquad (1)$$

Without restricting the generality, we can suppose that the connection matrix $\mathbf{J} = (J_{ij})_1^N$ is a symmetric one with zero elements on the main diagonal: $J_{ij} = J_{ji}$, $J_{ii} = 0$. We make use of physical terminology [1]. In what follows the binary variables $s_i = \{\pm 1\}$ will be called *spins*, $N$-dimensional vectors $\mathbf{s} = (s_1, s_2, ..., s_N)$ will be called *configuration vectors* or *configurations*, and the characteristic $E(\mathbf{s})$, which has to be minimized will be called *the energy* of the state $\mathbf{s}$. The commonly known procedure of minimization of the functional (1) is as follows: one randomly searches through $N$ spins, and to each spin the sign of the local field acting on this spin is assigned:

$$h_i(t) = \sum_{j=1}^{N} J_{ij} s_j(t). \qquad (2)$$

In other words, if the current state of the spin $s_i(t)$ coincides with the sign of $h_i(t)$ (if the spin is *satisfied*: $s_i(t)h_i(t) \geq 0$), one does not change the value of the spin: $s_i(t+1) = s_i(t)$; but if $s_i(t)$ and $h_i(t)$ are of opposite signs (if the spin is *unsatisfied*: $s_i(t)h_i(t) < 0$), in the next moment of the time one turns the spin over: $s_i(t+1) = -s_i(t)$.

It is well known that the energy of the state decreases when an unsatisfied spin turns over: $E(\mathbf{s}(t)) > E(\mathbf{s}(t+1))$. Sooner or later the system finds itself in the state, which is an energy minimum (may be this is a local minimum). In this state all the spins are satisfied and evolution of the system ends. In what follows the aforementioned algorithm will be called the <u>random dynamics</u> (in the theory of neural networks it is called the *asynchronous dynamics* [2]).

Another well known minimization procedure is the *synchronous dynamics*, when each time all the unsatisfied spins turn over simultaneously [2]. This approach is rarely used in minimization problems. First, this is because in this case it is impossible to guarantee the monotonous decrease of the functional $E(\mathbf{s})$. Second, synchronous dynamics is characterized by the presence of limit cycles of the length 2. Due to limit cycles the basin of attraction of local minima decreases.

In the papers [3], [4] a generalization of random dynamics, *the domain dynamics*, was proposed. The essence of the generalization is joining the spins together in larger blocks or, as they were called by the authors, *domains*. During the evolution not single spin turns over, but the whole block/domain. As for the rest the domain approach is the same as the standard random dynamics. It was found that the domain dynamics offers a lot of advantages comparing with other dynamic approaches. The domain dynamics is $k^2$ times quicker than the random dynamics, where $k$ is the averaged length of a domain (the number of spins joined in one block). Moreover, when using the domain dynamics, the energy $E(\mathbf{s})$ decreases monotonically, and it does not lead to limit cycles. Computer simulations with random Hebbian matrices [4] showed that the domain dynamics allowed one to obtain deeper local minima of the functional (1) than the standard random dynamics. However, there is an open question: how one has to choose the domains to obtain the deepest minima? In the paper we discuss this question.

## 2. DOMAIN DYNAMICS

For simplicity of presentation all definitions are given for the case when the first $k$ spins are joined in the 1st domain, the $k$ following spins are joined in the 2nd domain and so on. The last $k$ spins are joined in the last $n$th domain:

$$\mathbf{s} = (\underbrace{s_1,...,s_k}_{\text{1th domen}}, \underbrace{s_{k+1},...,s_{2k}}_{\text{2nd domen}},..., \underbrace{s_{(n-1)k+1},...,s_N}_{n\text{th domen}}). \tag{3}$$

In Eq. (3) $N = kn$ and $\mathbf{s}$ is an arbitrary configuration.

The total action onto the 1st domain from all other domains is the superposition of all interactions of the spins, which do not belong to the 1st domain, with the spins of the 1st domain. Then *the local domain field* acting onto the $i$th spin belonging to the 1st domain is equal to

$$h_i^{(d)}(t) = \sum_{j=k+1}^{N} J_{ij} s_j(t) = h_i(t) - \sum_{j=1}^{k} J_{ij} s_j(t), \ \forall\, i \leq k, \tag{4}$$

where $h_i(t)$ is the local field (2). In other words, the local domain field acting on the $i$th spin is obtained by means of elimination of the influence of the spins belonging to the same domain from the local field $h_i(t)$ (2). It is clear that the energy of interaction of the first domain with all other domains is equal to

$$E_1(t) = -F_1(t) = -\sum_{i=1}^{k} s_i(t) h_i^{(d)}(t).$$

In the same way we define the local domain field acting onto the spins belonging to the $l$th domain,

$$h_i^{(d)}(t) = h_i(t) - \sum_{j=(l-1)k+1}^{lk} J_{ij} s_j(t), \ \forall\, (l-1)k+1 \leq i \leq lk, \ l \in [2,n].$$

The energy of interaction of the $l$th domain with other domains is:

$$E_l(t) = -F_l(t) = -\sum_{i=(l-1)k+1}^{lk} s_i(t) h_i^{(d)}(t), \quad l = 2,..,n.$$

Then *the domain energy of the state* $\mathbf{s}(t)$ is

$$E^{(d)}(t) = \sum_{l=1}^{n} E_l(t) = -\sum_{l=1}^{n} F_l(t). \tag{5}$$

The energy (1), which we have to minimize, differs from the domain energy by the sum of the quantities $E_l^{(in)}$ that characterize the inter-domain interactions between spins only:

$$E(t) = E^{(d)}(t) + \sum_{l=1}^{n} E_l^{(in)}(t) = E^{(d)}(t) - \sum_{l=1}^{n} \sum_{i=(l-1)k+1}^{lk} \sum_{j=(l-1)k+1}^{lk} J_{ij} s_i(t) s_j(t), \qquad (6)$$

After that we can define *the domain dynamics* [3],[4]: one randomly searches through $n$ domains; if for the $l$th domain the inequality $F_l(t) \geq 0$ is fulfilled, the domain remains unchanged; but if $F_l(t) < 0$, in the next moment one turns over the $l$th domain – all the spins of the domain receive the opposite sings simultaneously:

$$s_i(t+1) = -s_i(t), \ i \in [(l-1)k+1, lk].$$

From Eq. (5) it is easy to obtain that when using the domain dynamics, the domain energy of the state decreases monotonically: if in the moment $t$ the $m$th domain is turned over, the domain energy goes down by the value $4|F_m(t)|$: $E^{(d)}(t+1) = E^{(d)}(t) - 4|F_m(t)|$. Note, in the same time the energy $E(\mathbf{s}(t))$ (1) that has to be minimized, goes down just by the same value. This follows from Eq. (6) and the obvious fact that simultaneous change of the signs of all spins belonging to the $m$th domain does not change the inter-domain energy $E_m^{(in)}$.

As a result of the aforementioned procedure sooner or later the system finds itself in *the domain* local minimum. In this state for all the domains the inequality $F_l(t) \geq 0$ is fulfilled, and the domain evolution ends. However, the local domain minimum is not necessarily a minimum of the functional (1). That is why in this state one has "to defrost" the domains and use the standard random dynamics. The dynamic system has the possibility to descend deeper into the minimum of the functional (1).

The consecutive use of the domain dynamics and than the standard random dynamics is based on the following argumentation. Minimization of the functional (1) can be interpreted as the sliding of the dynamic system down a hillside that is dug up by shallow local minima. In the case of the random dynamics two consecutive states of the system differ by an opposite sign of only <u>one</u> binary coordinate, namely those that is turned over during the given step of evolution. It is evident that under this dynamics the system sticks in the first occurring local minimum. On the contrary, under the domain dynamics two consecutive states of the system differ by opposite signs of <u>some</u> spin coordinates at once. The domain dynamics can be likening to sliding down by more "large-scale steps". It can be expected that such way of motion allows the system to leave aside a lot of shallow local minima, where it can stick in the case of random dynamics.

As noted above, for simplicity we gave all the definitions supposing that all the domains are of the same length $k$, and that spins joined in a domain have consecutive numbers – see Eq.(3). It is evident that in the general case domains can be of different lengths $k_l$, and any spins can be joined in a domain. In this connection there are some questions: does the choice of the domains influence the results of minimization? How the domains have to be organized? Have they been invariable for a given matrix **J**, or have they been organized randomly at every step of evolution? Generally, are there any argumentations, which can help? In the next section we formulate our recipe of the domain formation and present arguments proving it. In Section 4 we show the results of computer simulation.

**Note.** The idea of the domain dynamics is so natural that, probably, it was proposed more than once by different authors. In particular, during preparing this publication we found out that the domain dynamics is practically equivalent to the *block-sequential dynamics* presented in [5]. We note that in [5] the block-sequential dynamics was analyzed with regard to the problem of increasing the storage capacity of the Hopfield model. As far as we know, in [3], [4] the domain dynamics was used for minimization of the functional (1) for the first time.

## 3. CLUSTER APPROACH OF DOMAIN FORMATION

**1.** For the Hopfield model it is known the situation when domains are naturally appeared due to specific properties of the Hebbian connection matrix **J** [6]. Let us cite the corresponding results, accentuating the points we need. In the end of this Section we formulate the recipe of the domain formation in the general case.

Let us have $N$ $M$-dimensional vector-columns with binary coordinates $\mathbf{x}_i \in \mathbf{R}^M$, $x_i^{(\mu)} = \pm 1$, $i = 1,...,N$, $\mu = 1,..,M$. Vector-columns $\mathbf{x}_i$ are numerated by subscripts $i \in [1, N]$, and their coordinates by superscripts $\mu \in [1, M]$. The relation between the dimension $M$ of the vector-columns and their number $N$ does not mean for our purpose.

Let us construct ($M$x$N$)-matrix **X**, whose columns are $M$-dimensional vectors $\mathbf{x}_i$:

$$\mathbf{X} = \begin{pmatrix} x_1^{(1)} & x_2^{(1)} & ... & x_N^{(1)} \\ x_1^{(2)} & x_2^{(2)} & ... & x_N^{(2)} \\ ... & ... & ... & ... \\ x_1^{(M)} & x_2^{(M)} & ... & x_N^{(M)} \end{pmatrix}.$$

In the theory of neural networks **X** is called the pattern matrix. It is used to construct the Hebbian matrix that is ($N$x$N$)-matrix of scalar products of the vector-columns $\mathbf{x}_i$ (and its diagonal elements are supposed to be zeros):

$$J_{ij} = \frac{(1-\delta_{ij})}{M}(\mathbf{x}_i, \mathbf{x}_j) = \frac{(1-\delta_{ij})}{M}\sum_{\mu=1}^{M} x_i^{(\mu)} x_j^{(\mu)}, \ 1 \le i, j \le N.$$

Since the lengths of the vector-columns $\mathbf{x}_i$ are equal to $\sqrt{M}$, the matrix elements of $J_{ij}$ are cosines of the angles between the corresponding vectors: $|J_{ij}| \le 1$.

We shall examine a special case, when the number of <u>different</u> vector-columns in the matrix **X** is not $N$, but a smaller number $n$: $\mathbf{x}_1... \ne ..\mathbf{x}_l... \ne ..\mathbf{x}_n$, $n < N$. Then each vector-column $\mathbf{x}_l$ will be repeated some times. Let $k_l$ be the number of repetitions of the vector-column $\mathbf{x}_l$: $\sum_1^n k_l = N$. (The subscripts $l$, $m$ are used to numerate <u>different</u> vector-columns and related characteristics.) Without loss of generality it can be assumed that the first $k_1$ vector-columns of the matrix **X** coincide with each other, the next $k_2$ vector-columns coincide with each other, and so on; the last $k_n$ vector-columns coincide with each other too:

$$\mathbf{X} = \begin{pmatrix} \underbrace{x_1^{(1)} \ ... \ x_1^{(1)}}_{k_1} & \underbrace{x_2^{(1)} \ ... \ x_2^{(1)}}_{k_2} & ... & \underbrace{x_n^{(1)} \ ... \ x_n^{(1)}}_{k_n} \\ x_1^{(2)} \ ... \ x_1^{(2)} & x_2^{(2)} \ ... \ x_2^{(2)} & ... & x_n^{(2)} \ ... \ x_n^{(2)} \\ ... \ ... \ ... & ... \ ... \ ... & ... & ... \ ... \ ... \\ x_1^{(M)} \ ... \ x_1^{(M)} & x_2^{(M)} \ ... \ x_2^{(M)} & ... & x_n^{(M)} \ ... \ x_n^{(M)} \end{pmatrix}, \ k_l \ge 1. \tag{7}$$

The corresponding Hebbian matrix consists of $n \times n$ blocks. The dimensionality of the ($lm$)th block is equal to $(k_l \times k_m)$, and all its elements are equal to the same number $J_{lm}$ that is the cosine of the angle between the vectors $\mathbf{x}_l$ and $\mathbf{x}_m$ ($l, m = 1,...,n$). On the main diagonal there are quadratic $(k_l \times k_l)$-blocks; these blocks consist of ones only (with the exception of the diagonal, where all the elements are equal to zero):

$$\mathbf{J} = \begin{pmatrix} \overbrace{\begin{matrix} 0 & 1 & 1 \\ 1 & \ddots & 1 \\ 1 & 1 & 0 \end{matrix}}^{k_1} & \overbrace{\begin{matrix} J_{12} & \dots & J_{12} \\ \vdots & J_{12} & \vdots \\ J_{12} & \dots & J_{12} \end{matrix}}^{k_2} & \dots & \overbrace{\begin{matrix} J_{1n} & \dots & J_{1n} \\ \vdots & J_{1n} & \vdots \\ J_{1n} & \dots & J_{1n} \end{matrix}}^{k_n} \\ \begin{matrix} J_{21} & \dots & J_{21} \\ \vdots & J_{21} & \vdots \\ J_{21} & \dots & J_{21} \end{matrix} & \begin{matrix} 0 & 1 & 1 \\ 1 & \ddots & 1 \\ 1 & 1 & 0 \end{matrix} & \dots & \begin{matrix} J_{2n} & \dots & J_{2n} \\ \vdots & J_{2n} & \vdots \\ J_{2n} & \dots & J_{2n} \end{matrix} \\ \vdots & \vdots & \ddots & \vdots \\ \begin{matrix} J_{n1} & \dots & J_{n1} \\ \vdots & J_{n1} & \vdots \\ J_{n1} & \dots & J_{n1} \end{matrix} & \begin{matrix} J_{n2} & \dots & J_{n2} \\ \vdots & J_{n2} & \vdots \\ J_{n2} & \dots & J_{n2} \end{matrix} & \dots & \begin{matrix} 0 & 1 & 1 \\ 1 & \ddots & 1 \\ 1 & 1 & 0 \end{matrix} \end{pmatrix}, \; |J_{lm}| < 1, \; l, m \in [1, n]. \tag{8}$$

It was found out [6] that such organization of the connection matrix imposes the form of the local minima of the functional (1), namely: the local minimum certainly has the block-constant form

$$\mathbf{s} = (\underbrace{s_1 \; \dots \; s_1}_{k_1}, \underbrace{s_2 \; \dots \; s_2}_{k_2}, \dots, \underbrace{s_n \; \dots \; s_n}_{k_n}), \; s_l = \pm 1, \; l = 1, \dots, n. \tag{9}$$

In other words, the first $k_1$ coordinates of the local minimum have to be identical; the next $k_2$ coordinates have to be equal to each other, and so on. In Eq.(9) fragmentation into blocks of constant signs is defined by the block structure of the Hebbian matrix (or, and that is the same, by the structure of the repeating columns in the pattern matrix $\mathbf{X}$ (7)). The proof that the local minima have the block-constant form (9) is based on the fact that spins from one block have the same connections with all other spins, and under the action of the local fields they behave in the same way (see item 1 of Appendix).

Not all configurations of the form (9) are the local minima of the functional (1), but it is necessary to look for the local minima among these configurations only. In other words, for minimization of the functional (1), it is senseless to examine configurations with different signs inside blocks of constant signs. It makes sense to examine configurations of the form (9) only. All spins from the same block are satisfied or not satisfied simultaneously. In the last case it is senseless to turn over unsatisfied spins from the block separately, because this leads to a nonsensical configuration for which inside a block of constant sign there are coordinates with different signs. It turned out, that in this case it is possible to turn over the whole unsatisfied block simultaneously. This leads to a decrease of the energy (1). In other words, here the blocks of the constant signs play the role of <u>natural</u> domains.

**2.** Generally speaking, in this case the domains can be constructed in a random way also, and after that they can be used for the domain minimization of the functional (1). In the experimental part of the present work we tried to find out which way of the domain formation leaded to better results. However, before describing the experimental results, we would like to formulate the general rule of "the correct domains" formation:

**<u>Firstly</u>, a domain consists of spins whose inter-connections are stronger, then their connections with other spins.**

   **<u>Secondly</u>, the values of spins belonging to the same domain have to be equal.**

Due to evident relation of the general rule to the well known problem of the clustering of the symmetric matrix [7]-[9], this recipe will be called *the cluster principle* of the domains formation. In the clustering problem it is necessary to transform the symmetric matrix to a block-diagonal form, so that the matrix elements inside diagonal blocks are greater than the elements outside the diagonal blocks. In the end of this paper we discuss this problem in details.

Justification of the cluster principle is rather obvious. The matrix element $J_{ij}$ is considered as a measure of the connection between $i$th and $j$th spins. The validity of the cluster principle is based on the fact that strongly connected spins have to interact with the rest of the spins similarly. Therefore, we have a good chance that under action of an external field the strongly connected spins will behave themselves similarly.

Concluding this Section, let us note that the above constructed Hebbian matrix relates to a rather idealized situation, when the diagonal blocks of the matrix consist of ones. According to Eq.(9), when constructing "correct domains" we combine into one domain those spins that are <u>extremely</u> strong connected with each other. In practice such idealized situation can be found very rarely. In our computer simulations we examined not this idealized case only, but more realistic Hebbian matrices too.

## 4. RESULTS OF COMPUTER SIMULATION

**1.** In first series of the computer experiments we used the Hebbian matrices with extremely strong inter-group connections (see the item 1 of the previous Section). The external parameters of the problem were as follows: the dimensionality of the problem was $N=1000$, the number of patterns was $M=60$, the number of domains was $n=40$, sizes of domains $k_l$ were random numbers from the interval [1, 45]: $\sum_{l=1}^{40} k_l = 1000$; the coordinates $x_l^{(\mu)}$ in the matrix (7) took on the values $\pm 1$ equiprobable.

On the main diagonal of the Hebbian matrix (8) there were 40 $(k_l \times k_l)$-blocks from the ones only: $J_{ll}^{(in)} = 1$. Matrix elements outside these blocks were random quantities with the mean values equal to 0 and the dispersions equal to $1/M$: $<J_{lm}^{(out)}> = 0$, $\sigma(J_{lm}^{(out)}) = 1/\sqrt{M}$. The cluster principle of the domains formation provides us with 40 domains of the type (9) generated by 40 groups of strongly connected spins.

Altogether 200 such a matrices **J** were generated, and for each matrix the functional (1) was minimized using 3 dynamic approaches:

**1) RANDOM:** the standard random dynamics was set going from 1000 <u>random</u> start configurations;

**2) Random Domains (DM-RND)**: the domain dynamics was set going from the same random configurations with $n=40$ <u>random</u> domains of the identical size $k=25$ (25 spins with random numbers were included in a domain; inside a domain the spins had the values $\pm 1$ equiprobable);

**3) Cluster Domains (DM-CLS)**: the domain dynamics was set going from 1000 random start configurations of block-constant form (9).

When a domain local minimum was achieved, the domains were "defrosted" and under the standard random dynamics the system went down to a deeper local minimum of the functional (1). (Note, in the case of cluster domains the domain local minima are local minima of the functional (1) too - see the item 2 of Appendix; in this particular case the domain defrosting does not allow the system to go down deeper.)

Thus, for each matrix we obtained 3000 local minima. Then we found the deepest among them and calculated the frequency of the deepest minimum determination for each of the three dynamics: RANDOM, DM-RND and DM-CLS. The dynamics, for which the frequency of the deepest local minimum determination is the largest, has to be declared the best.

In Fig.1 the results averaged over 200 random tests are shown for all the three dynamics. Along the abscissa axis the three dynamics are marked off, along the ordinate axis we show the averaged frequency of the deepest minimum determination (in percentage terms). It is seen that in average the domain dynamics with cluster domains (DM-CLS) leads to the deepest minimum more then **30 times frequently** than the random domain dynamics (DM-RND) or the standard random dynamics (RANDOM). Here the preference of the cluster domain formation approach is evident.

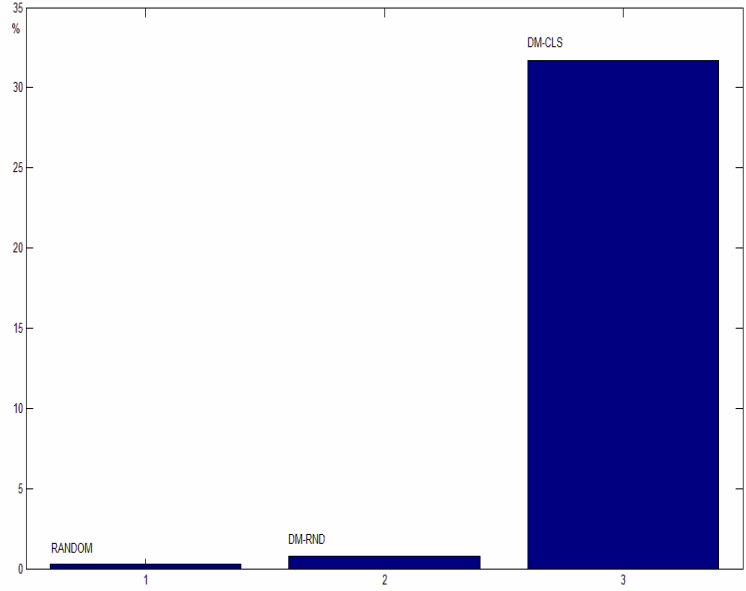

Fig.1. The average frequency of the deepest minimum determination for all the three dynamics.

**2.** In the described experiments the connections between spins inside "the correct" groups were equal to maximal value, because all $k_l$ vector-columns $\mathbf{x}_l$ in one group were the same. What happens if these groups of vectors are slightly "diluted": If the vector-columns inside "the correct" groups are not identical, but differ slightly? How this affects the result of minimization?

The aforementioned experiments were repeated for "diluted" groups of vector-columns $\mathbf{x}_l$. In this case the first $k_1$ vector-columns of the matrix (7) were not identical, but they were obtained as a result of multiplicative distortion of the vector $\mathbf{x}_1$: with the probability $b$ the coordinates of the vector $\mathbf{x}_1$ (independently and randomly) were multiplied by -1. Analogously, the next $k_2$ vector-columns of the matrix $\mathbf{X}$ were obtained by multiplicative distortion of random vector $\mathbf{x}_2$, and so on. The last $k_n$ vector-columns of the matrix $\mathbf{X}$ were the result of multiplicative distortion of random vector $\mathbf{x}_n$.

Then the connections between spins inside "the correct" groups were random quantities with the mean values $<J_{ij}^{(in)}> = (1-2b)^2$. Thus, the distortion probability $b$ characterized the level of inhomogeneity of "the correct" spin groups: the larger $b$, the less the mean value of the inter-group connection, the greater the inhomogeneity of "the correct" group of spins. (As before the estimate of the connections between spins from different groups was $J_{ij}^{(out)} \sim \pm 1/\sqrt{M}$.)

The values of $b$ and corresponding mean values of the inter-group connections are given in Table 1. Note, only for $b$=0.02 and $b$=0.05 "the correct" groups of spins can be regarded as strongly connected. Indeed, in these cases the mean values of the inter-group connections are $<J_{ij}^{(in)}> \approx 0.9$ and $<J_{ij}^{(in)}> \approx 0.8$, respectively. In other words, the angles between $M$-dimensional vectors $\mathbf{x}_i$ from "the correct" groups are less than $45°$. Such groups of vectors still can be regarded as compact, and the relative spins can be regarded as strongly connected. However, already for $b$=0.1 we have $<J_{ij}^{(in)}> = 0.64$, and for $b$=0.2 the mean value of the inter-group connection becomes entirely small: $<J_{ij}^{(in)}> = 0.36$. For such mean values of the matrix elements, the groups we mechanically continue to consider as "correct" groups, in fact are aggregations of slightly connected spin sub-groups. The spins inside these sub-groups can be strongly connected, but in the same time the sub-groups are connected rather slightly. Note, when combining some slightly connected sub-groups into one large group, in fact we organize random domains. One might expect that the more $b$, the more the results for DM-CLS-dynamics resemble the results for DM-RND-dynamics.

**Table 1.** The mean value $<J_{ij}^{(in)}>$ of the inter-group connection between spins as function of the distortion level $b$.

| $b$ | 0.02 | 0.05 | 0.1 | 0.2 |
|---|---|---|---|---|
| $<J_{ij}^{(in)}>$ | 0.92 | 0.81 | 0.64 | 0.36 |

In Fig.2 for all three types of the dynamics it is shown how the mean frequency of the deepest minimum determination depends on the parameter $b$. We see that when $b$ increases, the results for the cluster domain dynamics (DM-CLS) progressively less differ from the results for the random domain dynamics (DM-RND), particularly beginning from $b$=0.1. It can be expected, since when $b$ increases, "the correct" groups of spins resemble the random domains progressively.

We pay ones attention that the results of the random domain dynamics are only slightly better than the results of the standard random dynamics: for all values of $b$ the DM-RND-plot is only slightly higher than the RANDOM-plot. Possible this can be explained by the fact that in our experiments the length of the random domains, $k = 25$, was far from optimal. In Ref. [4] the random domain dynamics has been examined for different lengths of domains. It was found out that the best results could be obtained when $k$ =2.

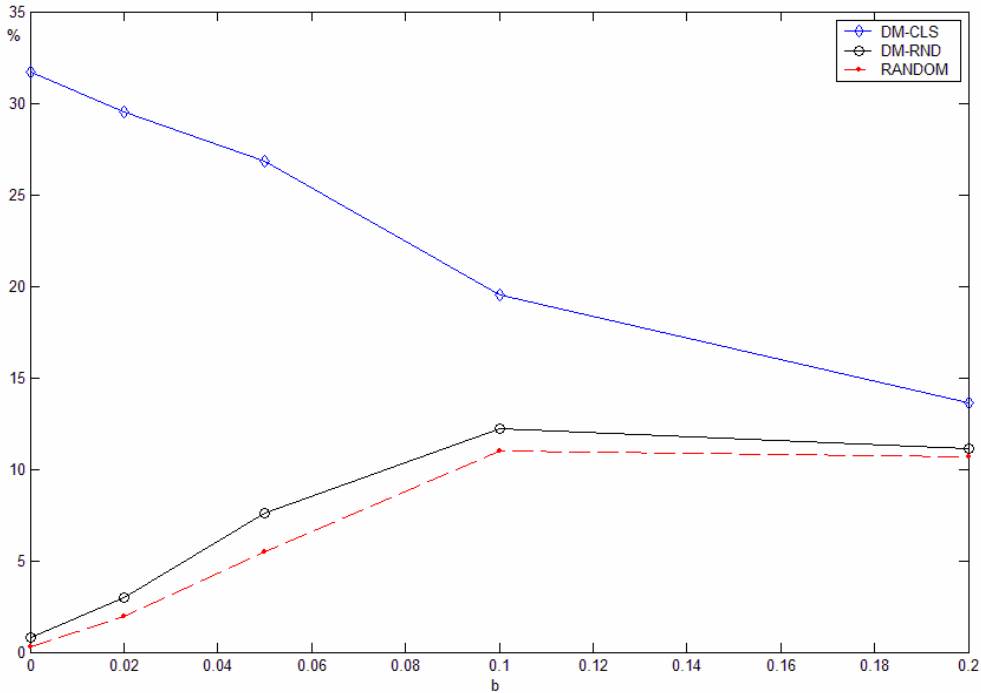

Fig.2. The mean frequency of the deepest minimum detecting as function of the distortion parameter $b$ for all the three dynamics.

**3.** After the dynamical system finds itself in the domain local minimum, one has to "defrost" the domains and to use the standard random dynamics. At that the system gets chance to decrease its energy still more getting into the local minimum of the functional (1). Suppose $D$ is the value of the <u>domain</u> local minimum, and $E$ is the depth of the <u>final</u> local minimum; it is evident that $D < 0$ and $E \leq D$. Then $d=D/E$ and $r=(E-D)/E$ are relative quantities characterizing which part of the depth of the local minimum is due to the domain dynamics, and which part is due to the random dynamics:

$$0 \leq d, r \leq 1, \quad d + r = 1.$$

If, for example, $r \approx 0$, we conclude that practically all the depth of the local minimum is defined just by the contribution of the domain dynamics, and some increase of the local minimum depth due to the random dynamics is negligible. On the other hand, if $r \approx 1$, the situation is reversed: the domain dynamics does not influence significantly on the local minimum depth, and the random dynamics plays the main role. In fact, $d$ and $r$ characterize the relative contribution of the domain and random dynamics to the depth of the local minimum. It is interesting to compare the values of $d$- and $r$-characteristics for both variants of the domain dynamics.

To do this, we, at first, for each matrix averaged the $r$-characteristics over 1000 random starts, and then we averaged it over 200 random matrices. This has been done for two examined types of the domain dynamics. The plots of the averaged $r$-characteristics are shown in Fig.3. Here the distortion level $b$ is along the abscissa axis, and $r$-characteristics for the cluster domain (DM-CLS) and random domain (DM-RND) dynamics (averaged over 200000 starts) are along axis of ordinates.

We see that random domains give us $r \approx 1$. In other words, when random domains are used in average only 5% of the depth of the local minimum is defined by the contribution of the domain dynamics. The remaining 95% of the depth of the local minimum accounts for the standard random dynamics. This result is practically independent of the distortion level $b$ (see the (DM-RND)-curve). The same as in the end of the previous item we can say that when random domains of the length $k=25$ are used the result of standard random minimization is improved only slightly.

Vice versa, $r \ll 1$ when cluster domains are used. In this case the largest part of the depth of the local minimum is defined by the contribution of the domain dynamics namely, and the contribution of the random dynamics is comparatively small (see the (DM-CLS)-curve). In the absence of distortions ($b=0$) the domain local minima are just the minima of the functional (1) (see item 2 of Appendix); so, in this case the strict equality $r = 0$ is fulfilled. As far as $b$ increases, the $r$-characteristics increases too. However, it happens rather slowly. The contribution of the domain dynamics predominates in the entire examined interval of $b$.

## 5. Conclusions

The obtained results are the evidence of the productivity of the cluster principle of domains formation. At that the number and the structure of domains are defined as a result of the connection matrix clustering. A transformation of a matrix to the block-diagonal form when its matrix elements inside the diagonal blocks are greater than outside these blocks is usually implied by matrix clustering. There are a lot of different approaches to solution of the problem. A review of methods of clustering can be found in [7]-[9].

The clusterization procedure proposed in [10] recommended itself rather good for the correlation type connection matrices. They are the matrices whose elements are scalar products of a set of vectors. During the time of the order of $O(N^2)$ this procedure allows one not merely to choose compact groups of vectors, but gain the understanding of which of these groups are closer and which are farther to each other. This information is useful for domains formation. Up to now we failed to generalize the clusterization procedure [10] onto connection matrices of the general (not obviously correlation) forms.

Author is grateful to Artem Murashkin, who has done computer simulations for this paper. The work was done in the framework of the project «Intellectual computer systems» (the program 2.45) under financial support of Russian Basic Research Foundation (grant 06-01-00109).

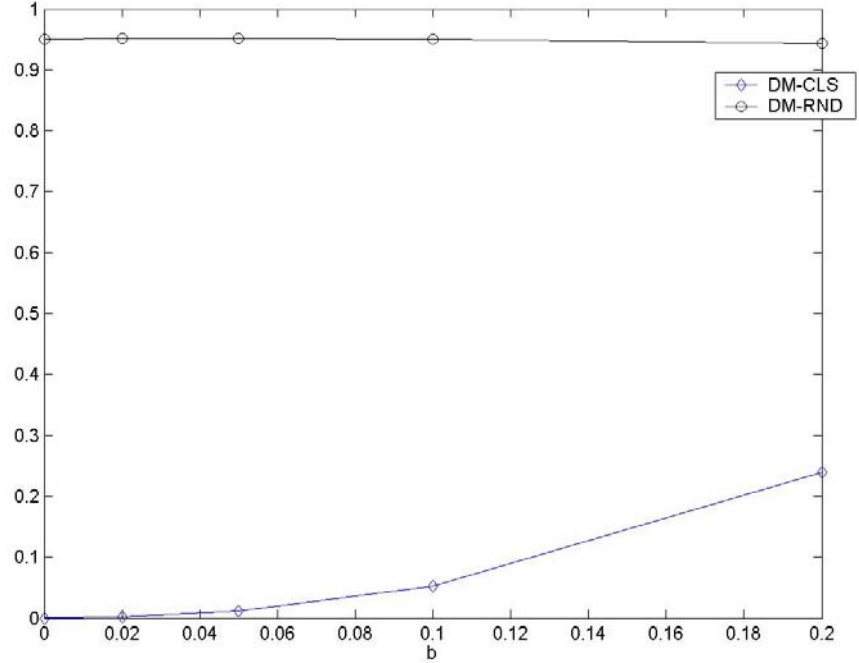

Fig.3. The averaged *r*-characteristics for the domain dynamics for cluster (DM-CLS) and random (DM-RND) domains.

## APPENDIX

**1.** Let us show that when the connection matrix consists of the blocks (8), the local minima of the functional (1) have the piecewise constant form (9). To do this it is sufficient to prove, for example, that the first $k_1$ coordinates of the local minimum have to be equal to each other. Argumentation that ascertains this statement will be carried out for the first two coordinates.

Suppose the configuration $\mathbf{s} = (s_1, s_2, s_3, ..., s_N)$ is a local minimum. Then each of the coordinates $s_i$ must have the same sign as the local field acting on this coordinate: $s_1 h_1 = s_1 \sum_{j=1}^{N} J_{1j} s_j \geq 0$ and $s_2 h_2 = s_2 \sum_{j=1}^{N} J_{2j} s_j \geq 0$. Since the matrix $\mathbf{J}$ has the form (8), the local fields can be written as: $h_1 = s_2 + H$ and $h_2 = s_1 + H$. For both cases the second term $H = \sum_{j=3}^{N} J_{ij} s_j$ is the same, since all matrix elements of the first and the second rows (with subscripts j>2) are the same: $J_{1j} = J_{2j}$, $j > 2$ (see Eq.(8)). Thus two inequalities have to be fulfilled simultaneously: $s_1 s_2 + s_1 H \geq 0$ and $s_2 s_1 + s_2 H \geq 0$. If we suggest that the coordinates $s_1$ and $s_2$ differ, then $s_1 s_2 = -1$, and as a result we obtain $1 \leq H$ and $1 \leq -H$. This is impossible. Consequently, our suggestion is incorrect, and the coordinates $s_1$ and $s_2$ must coincide. This completes the proof that the local minima of the functional (1) with the matrix **J** (8) has the piecewise constant form (9).

**2.** It is easy to see that for the block-constant Hebbian matrix (8) any configuration of the form (9) that is a <u>domain</u> local minimum is a local minimum of the functional (1) also. Indeed, suppose for the first domain the inequality $F_1 = \sum_{i=1}^{k_1} s_i h_i^{(d)} \geq 0$ is fulfilled. Since first $k_1$ coordinates $s_i$ (which just form the domain) are equal to each other, the sum in the right-hand side of the last expression is the sum of $k_1$ equal terms: $F_1 = k_1 s_i h_i^{(d)}$. Consequently, $s_1 h_1^{(d)} = s_1 \sum_{j=k+1}^{N} J_{1j} s_j \geq 0$. If we use the expression (4) connecting the local field $h_i$ with the domain local field $h_i^{(d)}$, we obtain that the following expression is also positive:

$$s_1 h_1 = s_1 h_1^{(d)} + s_1 \sum_{j=1}^{k_1} J_{1j} s_j = s_1 h_1^{(d)} + (k_1 - 1) > 0 . \qquad (10)$$

This means that the sign of the coordinate $s_1$ coincides with the sign of the local field $h_1$, acting on this coordinate from the entire network. In other words, in this case a domain local minimum is also a local minimum of the functional (1). It is interesting that the opposite is incorrect: from the piecewise constant form (9) being a local minimum of the functional (1) it does not follow that it is a domain local minimum also. It can be easily seen from the expression (10): $s_1 h_1^{(d)}$ can be negative, but if $k_1$ is sufficiently large, the sum in the right-hand side of Eq.(10) is positive. In other words, two inequalities are fulfilled simultaneously: $s_1 h_1 > 0$, $s_1 h_1^{(d)} < 0$. By the way, from the last statement it follows that the domain dynamics allows one to get out from shallow local minima of the functional (1).